Preprint version.

# Beyond travel mode: urban context shapes active mobility's mental health effects over time

Shujuan Chen[1,2,*], Yue Li[1,2], Ying Jin[1]

[1] Martin Centre for Architectural and Urban Studies, University of Cambridge, Cambridge, UK
[2] Cambridge Public Health, University of Cambridge, Cambridge, UK
* Correspondence to Shujuan Chen, The Martin Centre, Department of Architecture, University of Cambridge, 1 Scroope Terrace, Cambridge CB2 1PX, sc2343@cam.ac.uk

## Abstract

Active mobility is widely promoted for sustainable and healthier living, but whether it translates into equitable mental health benefits across individuals and places over time remains unknown. Using causal machine learning and causal deep learning in 264168 UK adults, we find substantial inequalities in individualized effects of active mobility on anxiety, depression, and common mental disorders. These inequalities widen over time and are strongly structured by urban context. For example, anxiety risk at follow-up ranges from a 40.6% reduction to a 10.1% increase across individuals, versus a 10.4% reduction to a 0.1% increase at baseline. Benefits are greatest in greener, safer, less polluted, and less deprived neighborhood environments, with 81.8% of individuals experiencing above-average benefits and mean anxiety risk reduced by 26.4%, versus 10.4% of individuals and 7.4% reduction in the least supportive environments. Urban compact form further modifies these effects through nonlinear interactions with neighborhood environments, amplifying benefits only under supportive conditions. Despite these strong environmental gradients, genetic moderation is negligible.

## Introduction

As cities expand and health challenges intensify, active mobility is increasingly recognized for its potential health co-benefits beyond decarbonization (*1-9*). Mental health is a particularly important domain for such benefits, given both the scale of the global burden, with one in seven individuals affected by a mental disorder and 71% not receiving treatment (*10*), and the potential for active mobility to serve as a routine behavioral intervention supporting psychological wellbeing on a daily basis (*11, 12*). Yet despite growing interest in promoting active mobility across sectors, evidence on its mental health effects remains unclear, particularly regarding who benefits and under what conditions. This lack of evidence on heterogeneity constrains the ability to design targeted, equitable interventions.

Emerging evidence suggests that health responses to active mobility, including psychological outcomes, may not be uniform. At the individual level, the average effects of active mobility on cardiovascular disease have been found to vary by sex and body composition (*7, 8*), while effects on depressive severity were greater among younger adults and women (*3*), Associations with dementia risk appear strongly modified by genetic predisposition, such as Apolipoprotein E ε4 carrier status (*13*). At the same time, the places in which active mobility occurs may matter. Some observational studies have reported lower psychological distress among active commuters, with stronger associations in greener and less deprived settings (*5, 14, 15*). Similarly, active commuting through natural environments has been linked to better health outcomes, with benefits moderated by environmental quality (*16*). In practice, planning frameworks worldwide further shape these contexts by promoting active mobility within compact urban development characterized by high density and accessibility (*1, 2*). Yet evidence on how compact urban form affect the health benefits of active mobility remained limited, as measures based only on population density do not capture compact urban form sufficiently (*17*).

Several challenges in the existing work must be addressed to predict the psychological effects of active mobility with precision. First, prior studies have typically examined urban features in isolation, limiting their ability to characterize the full complexity of urban contexts to which active travelers are exposed, including both neighborhood environments and urban form (*1*). Second, most have relied on single mental-health indicators at single time points, providing only a partial view of psychological wellbeing across time and the continuum from symptoms to clinical diagnosis. Third, existing work has relied on estimating average associations from conventional risk models, rather than predicting individual causal effects that compare outcomes under active mobility with the counterfactual of motorized travel. With these limitations, it remains unknown whether the mental health benefits of active mobility are equitably distributed, or concentrated among specific populations and places.

These gaps in evidence have direct policy consequences. If health co-benefits of active mobility are context-dependent, uniform promotion policies may fail to deliver them equitably and could widen health inequalities, especially as active mobility becomes more central to public health and urban strategies. Given the limited feasibility of randomized experiments for estimating causal effects of such behavioral interventions, addressing these gaps requires analytical tools that go beyond average effects to estimate individual treatment effects and examine how they are structured by urban context. Although some machine learning (ML) or deep learning approaches for heterogeneous treatment effect estimation have been proposed in precision health, they have rarely been applied to urban mobility and health research (*18*). To date, no existing framework integrates individualized causal inference with the multiple dimensions of urban context needed to inform place-based health and urban policy.

To address the gaps, we developed a causal framework incorporating stacked machine learning and deep survival modeling to quantify individualized and context-specific treatment effect of active mobility on mental health. Using prospective cohort data from 17977 UK Biobank neighborhoods over the period from 2006 to 2020, linked with environmental, electronic health records, and genetic data, we applied this framework to multiple mental health outcomes, including anxiety, depression, and common mental disorders (CMDs). We then investigated how these effects vary across two key dimensions of urban context: neighborhood environments and compact urban form. To characterize these dimensions, we constructed two corresponding metrics: a composite environmental supportiveness measure for transport health (THES), integrating physical, natural, and social conditions, and a functional urban compactness measure reflecting the intensity of urban activity and accessibility. We further examined how these two dimensions interact to shape mental health gains from active mobility promotion, identified thresholds and trade-offs in these effects, and assessed whether individual genetic susceptibility modifies them. These findings could provide evidence for targeted policies to advance health equity in densifying cities. Fig. 1 shows the data, models, and study design.

Preprint version.

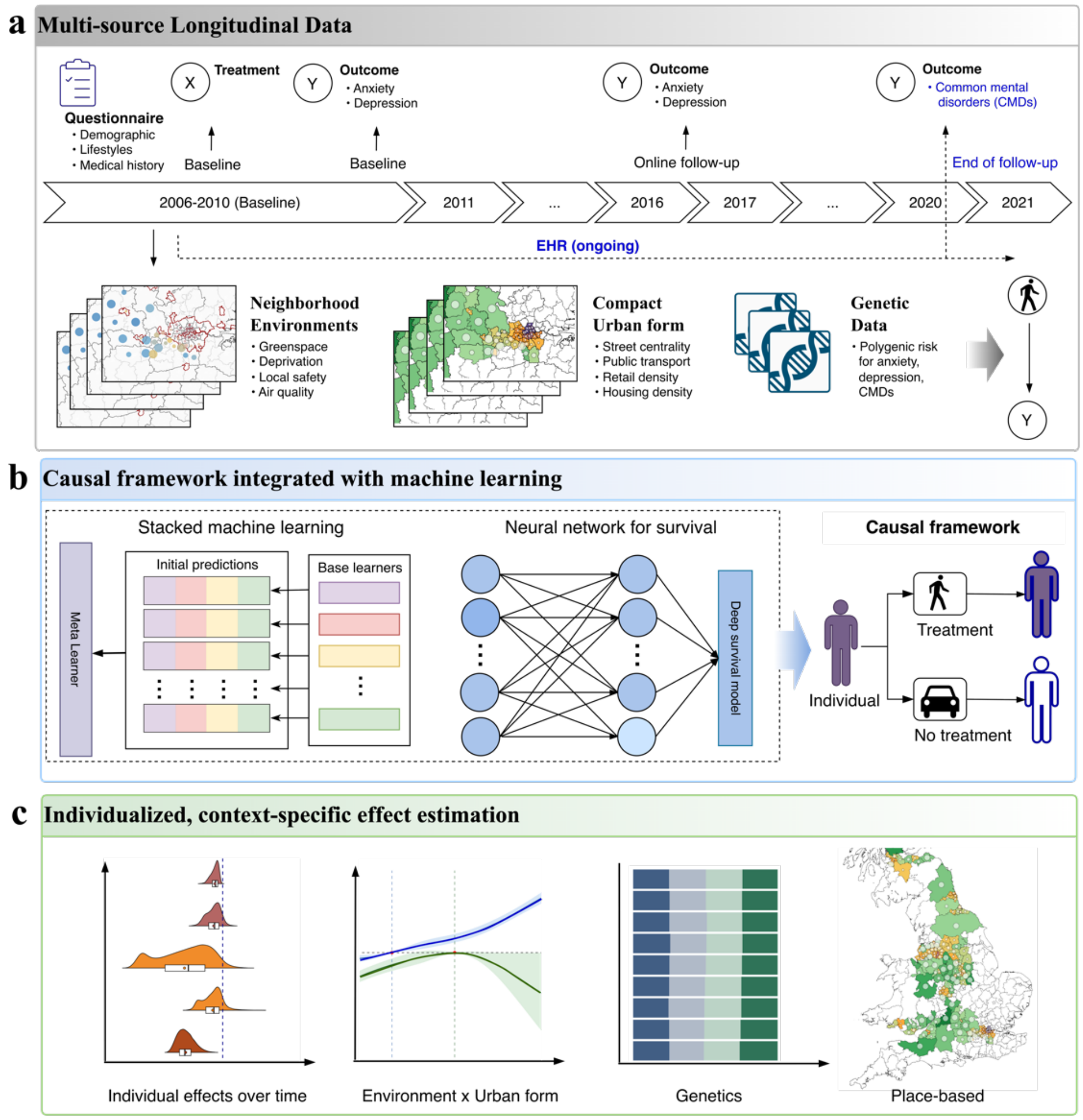

**Fig. 1. Data, methods, and study design for estimating individual-level effects of active mobility on multiple mental health outcomes over time across two dimensions of urban context: compact urban form and neighborhood environments. a.** Overview of the study data, derivation of the composite measures of compact urban form and neighborhood environments, and their modifying roles in shaping the relationship between mobility and mental wellbeing. **b.** Causal framework integrating machine learning for treatment-effect estimation. **c.** Estimation and inference of individualized and context-specific effects of active mobility.

## Results

This study included 264168 (52.6%) of the UK Biobank participants, aged 38–70 years (mean age: 52.6 years), from 17977 small-area neighborhoods with complete data on travel mode. At the baseline (2006–2010), 34747 participants screened positive for depression and 72762 participants screened positive for anxiety. The follow-up online mental health assessment in 2016 to 2017 was completed by 88915 participants, of whom 2434 screened positive for depression and 1993 screened positive for anxiety. Over a median follow-up of 10.8 years from baseline to 2020, 14273 participants received a first diagnosis of CMDs recorded in primary care or inpatient admissions. Participant characteristics are summarized in Supplementary Table S4.

We developed a neighborhood environmental supportiveness score (THES) integrating green space, air pollution, deprivation, and crime into a single measure ranging from 5 to 50, with higher values indicating more supportive environments. Urban compactness was constructed based on densities of public transport, street movement, retail, and residential activity, ranging from 4 to 40 with higher values indicating greater compactness (details in Materials and Methods).

Across the UK, environmental supportiveness was generally higher in northern and rural areas and lower in major urban centers, while urban compactness showed the reverse pattern, concentrated in London and other large cities (Fig. 2a, b). Environmental supportiveness decreased with increasing urban compactness, reflecting higher crime, lower green space, higher air pollution, and greater deprivation in more compact areas (Fig. 2c, d).

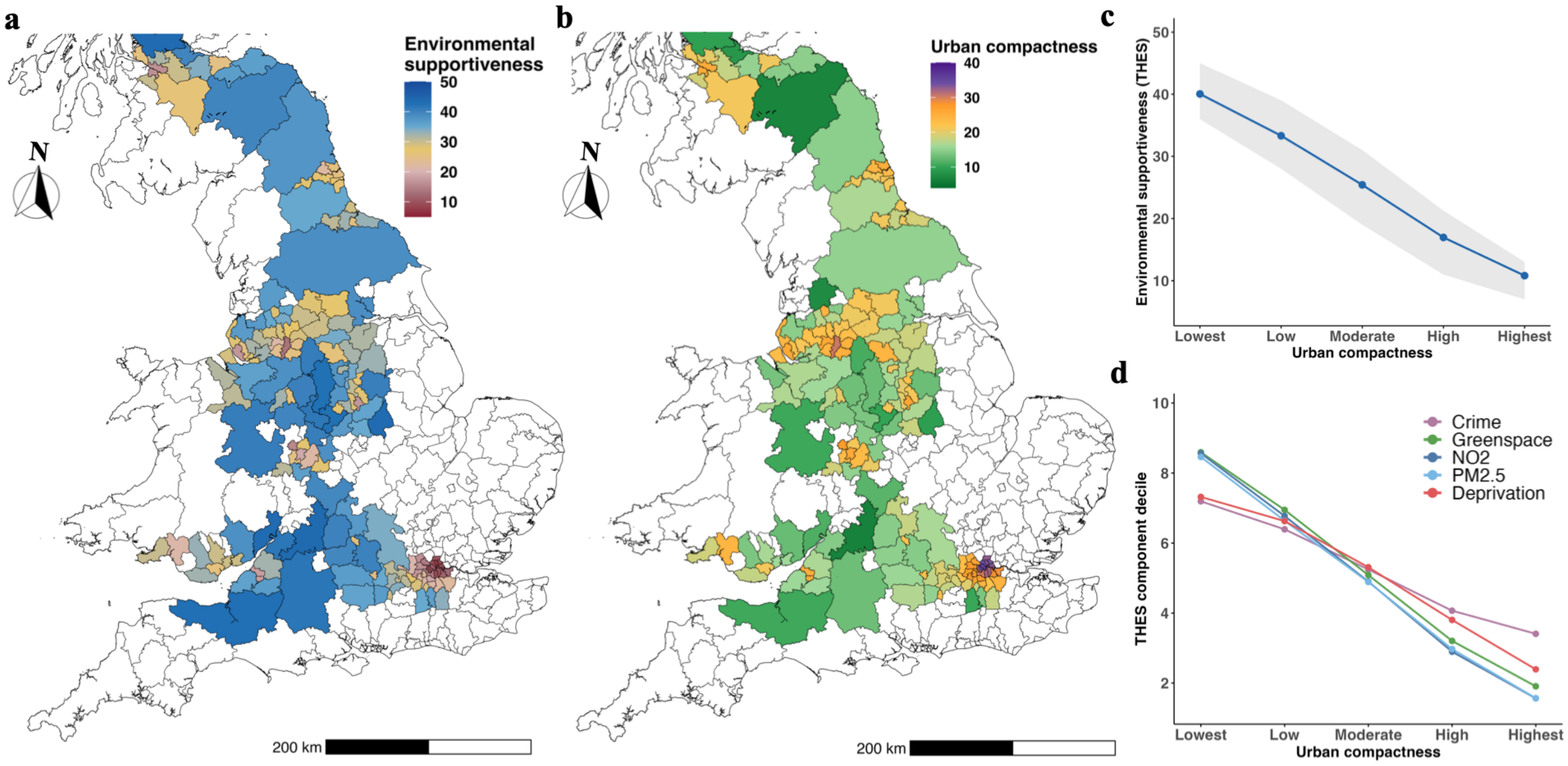

**Fig. 2. Descriptive patterns of neighborhood environmental supportiveness (THES) and functional urban compactness.** THES is a composite score based on green space, air pollution, deprivation, and crime, ranging from 5 to 50, with higher values indicating safer, greener, less polluted, and less deprived environments. Functional urban compactness is based on densities of public transport, street movement, retail, and residential activity, scored from 4 to 40, where higher values reflect greater compactness. **a**, Mean THES across local authority districts. **b**, Mean functional urban compactness across local authority districts. **c**, Mean THES across urban compactness groups, with shaded bands showing the 25th to 75th percentile range. **d**, Mean decile scores of the five THES components across the same compactness groups. Higher deciles indicate more supportive conditions in the direction shown.

## Individual treatment effects (ITEs)

Individualized treatment effects (ITEs) were expressed as relative risks (RR), the ratio of predicted outcome risk for each individual under active mobility to that under counterfactual motorized travel, where values below 1 indicate reduced risk.

Overall, active mobility was associated with highly unequal mental health benefits across individuals, and these inequalities were larger at follow-up than at baseline. On average, if individuals had taken active mobility instead of motorized travel, their risks of having mental disorders would have been 3.8% lower for depression and 3.3% lower for anxiety at baseline, 3.9% lower for depression and 15.6% lower for anxiety at follow-up, and 16% lower for 10-year risk of incident CMDs (Fig. 3).

Specifically, for the online follow-up outcomes, ITEs for active mobility ranged from 0.839 to 1.176 for depression (mean RR 0.961; $p > 0.05$), and from 0.594 to 1.101 for anxiety (mean RR 0.844; $p < 0.05$). For mixed mode, ITEs ranged from 0.939 to 1.189 for depression (mean RR 1.023; $p > 0.05$) and from 0.652 to 1.029 for anxiety (mean RR 0.853; $p < 0.05$). For clinically diagnosed CMDs from health records, ITEs ranged from 0.754 to 0.969 for active mobility (mean RR 0.838; $p < 0.05$), and from 0.867 to 1.056 for mixed mode (mean RR 0.936; $p < 0.05$; Fig.3).

For baseline outcomes, ITEs for active mobility ranged from 0.860 to 1.106 for depression (mean RR 0.962; $p < 0.05$) and similar trends for anxiety (ITE range: 0.896 to 1.001; mean RR 0.967, $p < 0.001$). Mixed-mode ITEs ranged from 0.933 to 1.076 for depression (mean RR 0.996; $p > 0.05$) and showed similar patterns for anxiety.

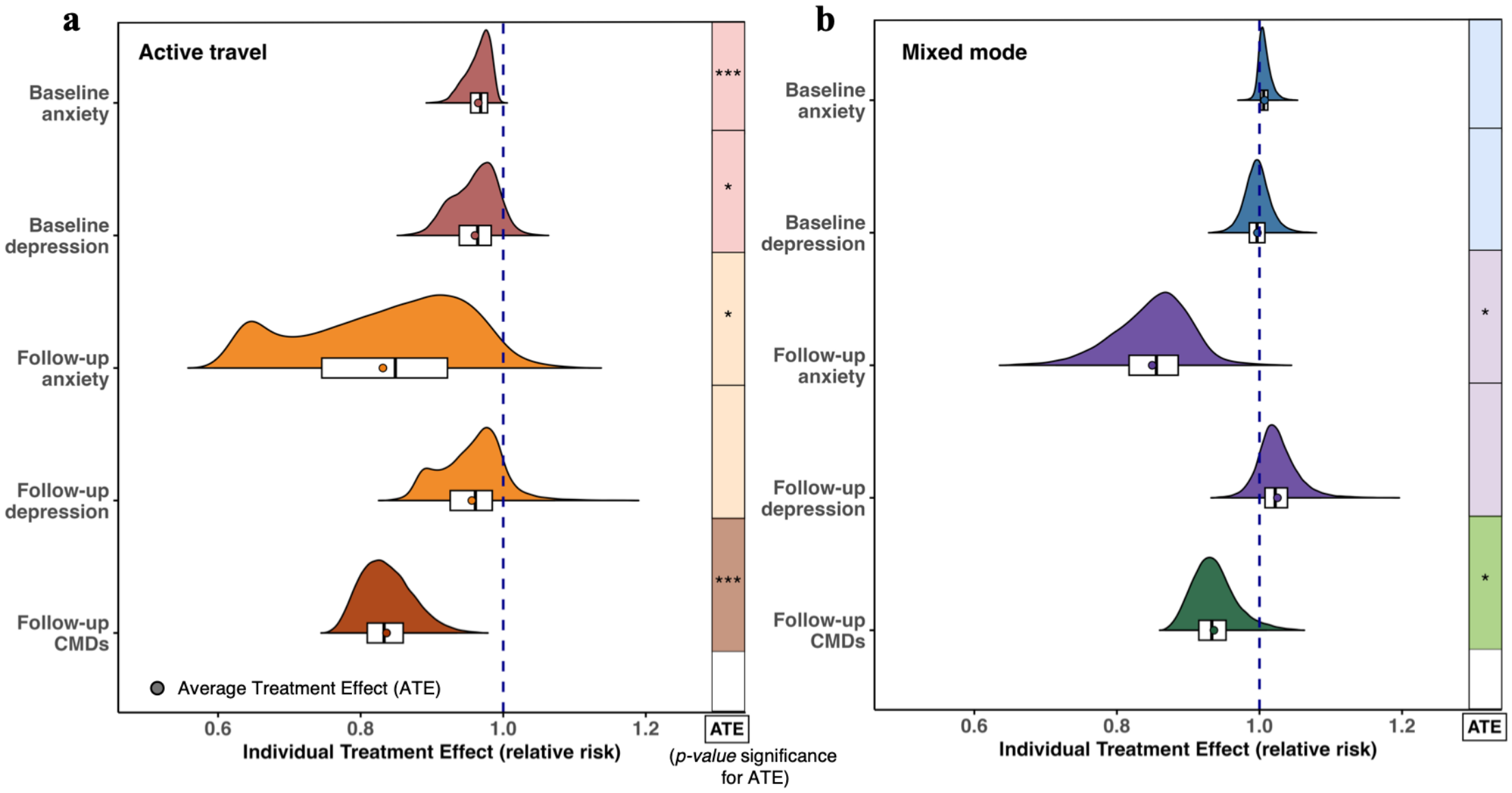

**Fig. 3. Distribution of individual treatment effects (ITEs) on different mental health outcomes across time. a.** ITEs distribution for active mobility. **b.** ITEs distribution for mixed-mode travel. Ridge plots show the ITE distribution. Box plots show interquartile range (bounds of boxes), median is indicated by center line, mean is indicated by dot. Vertical dash line at RR=1.0 indicates no effect. Significance for the ATE is the follows: *** $p < 0.001$, ** $p < 0.01$, * $p < 0.05$.

## ITEs, neighborhood environments, and urban compactness

### ITEs and neighborhood environments

The dose-response relationship between environmental supportiveness and ITEs of active mobility for follow-up anxiety was monotonic and nonlinear ($p < 0.0001$ for nonlinearity), with benefits accelerating markedly in more supportive environments (Fig. 4a-e). Across categorical levels of environmental supportiveness, ITE distributions differed significantly ($p < 0.0001$), with the largest benefits concentrated in the most supportive environments (Fig. 4f-j). In the highest environmental supportiveness category (45–50), 81.7% of people showed enhanced benefits (RR below the mean RR) under active versus motorized travel, whereas in the lowest category, 10.4% participants showed enhanced effects.

On average, active mobility in the most supportive environments corresponded to a markedly lower incident anxiety risk, with a 26.4% risk reduction (RR 0.736, 95% CI: 0.731–0.741). By contrast, in the least supportive environments the risk reduction was 7.4% (RR: 0.926, 95% CI:0.923–0.928). Overall, the mean risk reduction was 15.6%.

### Joint modification of neighborhood environments and urban compactness

The mental health benefits of active mobility varied with urban compactness, but this relationship depended on environmental supportiveness. In more supportive environments, the association was curvilinear, with a threshold at urban compactness scores of approximately 26 to 29. Below this threshold, benefits decreased with increasing compactness, whereas above it, they increased. In less supportive environments, benefits were generally weaker and declined monotonically as compactness increased. The interactions between environmental supportiveness and urban compactness were statistically significant ($p < 0.01$) and showed broadly consistent patterns across all health outcomes (Fig. 5; Supplementary Table S6). The underlying directed acyclic graph is shown in Fig. 5f.

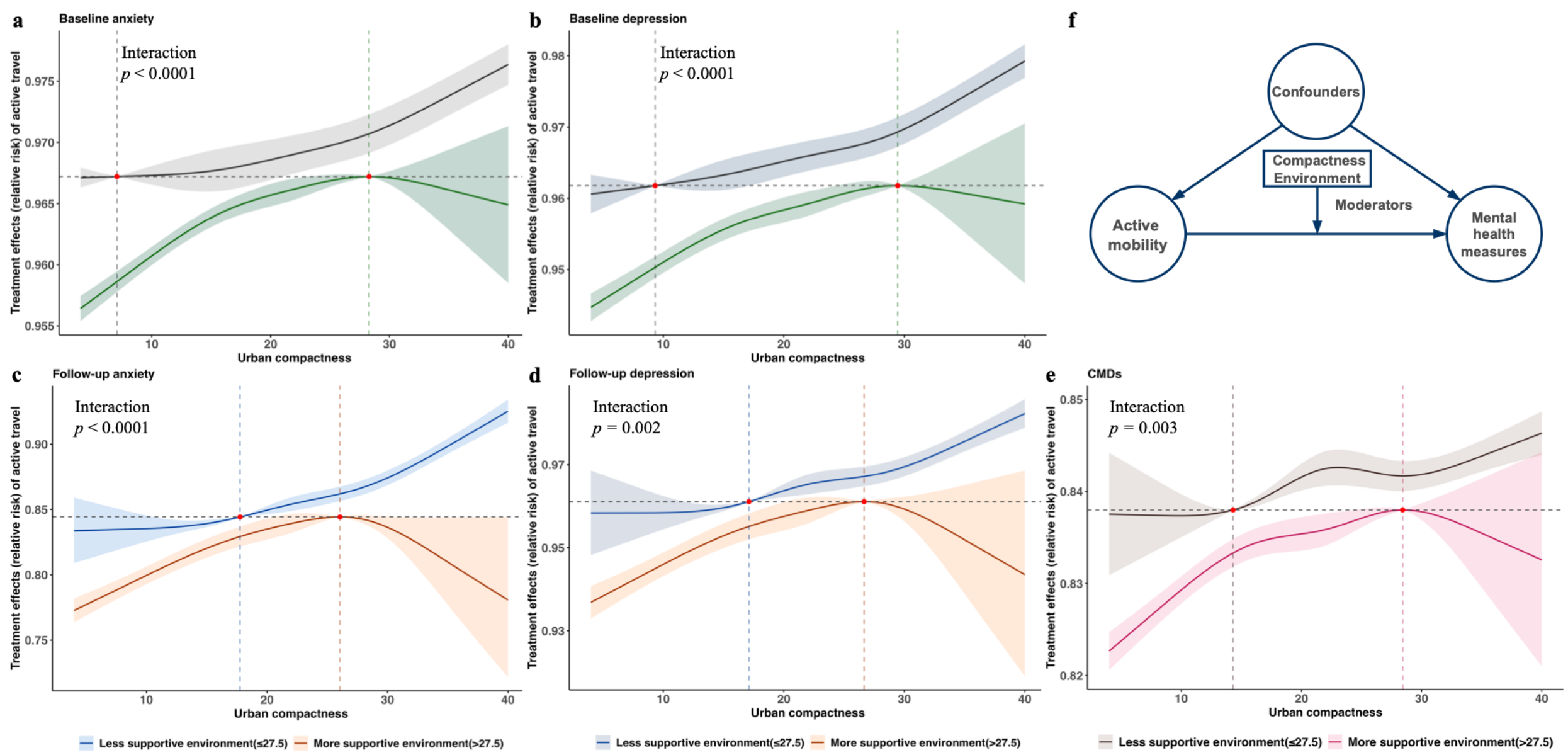

**Fig. 5. Dose-response associations between urban compactness and individualized treatment effects (ITEs), stratified by neighborhood environmental supportiveness.** Urban compactness modified the mental-health gains of active mobility through nonlinear interactions with neighborhood environments, with above-average benefits found only in more supportive settings and the weakest benefits at intermediate compactness. **a-e**, Associations for baseline anxiety, baseline depression, follow-up anxiety, follow-up depression, and incident common mental disorders (CMDs), respectively. **f**, Schematic of the moderation analysis linking urban compactness, environmental supportiveness, active mobility, and mental health outcomes. Solid lines show estimated associations and shaded bands indicate 95% confidence intervals. Interaction p values test whether association between ITEs with urban compactness differs across environmental supportiveness groups.

## ITEs and genetic susceptibility

Finally, we evaluated whether individual genetic susceptibility to mental disorders, characterized using polygenic risk score (PRS) that summarizes the cumulative effect from common genetic variants associated with each disorder, modified the mental health effects of active mobility, and whether it interacted with environmental supportiveness. In Fig. 8, the composition of ITE quartiles varied markedly across THES categories, but showed little variation across PRS levels. Although differences in ITE quartile composition across PRS strata was statistically significant for some outcomes, the effect sizes were negligible (Cramer's V 0.0103–0.0211, Supplementary Table S7ii). Consistently, PRS differed trivially between participants with high vs low ITEs (Cohen's d -0.008 – 0.03; Supplementary Table S7i, Fig. S8). We found no evidence of joint modification by genetic risk and environmental supportiveness ($p > 0.05$, Supplementary Table S7iii). Similar results were observed for baseline health outcomes (Supplementary Fig. S7).

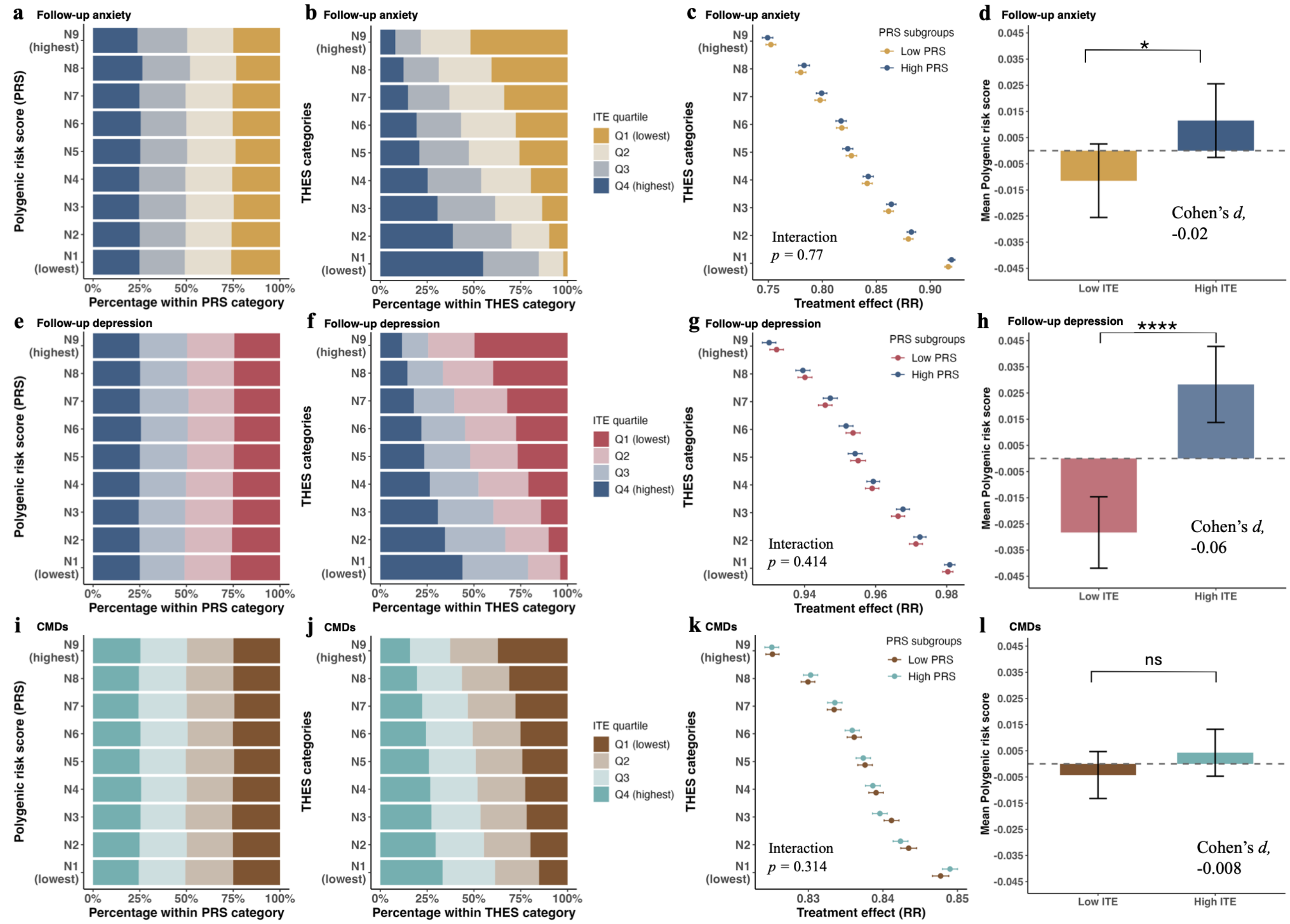

**Fig. 8. ITEs on mental health outcomes and moderation by genetic risk, compared to neighborhood environments.** Genetic moderation of the mental-health gains from active mobility was substantially weaker than moderation by neighborhood environments. **a, e, i,** Composition of ITE quartiles across polygenic risk score (PRS) categories (N1-N9, equal-sized groups) for the corresponding outcomes. **b, f, j,** Composition of ITE quartiles (Q1-Q4) across environmental supportiveness (THES) categories (N1-N9). **c, g, k,** Mean ITEs with 95% confidence intervals across THES categories, stratified by high versus low PRS (median split); p values test the PRS × THES interaction, assessing whether the THES-ITE association differs by genetic risk. **d, h, l,** Mean PRS among participants with high versus low ITEs (median split). Brackets indicate significance of group differences (**** $p < 0.0001$, *** $p < 0.001$, ** $p < 0.01$, * $p < 0.05$, ns not significant). Effect size is summarized using Cohen's d, calculated as the standardized mean PRS difference between high versus low ITE.

## Discussions

In this study, we estimated the individual mental health effects of active mobility across multiple time points, and assessed how these effects were structured by neighborhood environments, functional urban compactness, and genetic susceptibility. Drawing on a unified causal framework combining machine learning modeling, we identified the neighborhood environments and urban form profiles that enhance mental health gains of active mobility, and the extent to which genetic predisposition moderates them.

This study represents one of the first efforts to characterize the full distribution of individual-level health benefits of active mobility, and to compare these effects across multiple follow-up horizons. Active mobility showed beneficial treatment effects that varied considerably across individuals, with on average stronger benefits at follow-up than at baseline, particularly for anxiety and clinically diagnosed CMDs. Average risk reductions were larger for anxiety at follow-up (15.6% lower risk) and for 10-year incident CMDs (16.2% lower risk). These increasing benefits may reflect the physical activity (PA) involved, which can support neurocognitive adaptation, self-efficacy, and well-being that strengthen protective effects over time (*21, 22*). Risk reductions were larger for anxiety than depression, potentially because anxiety is more stress-reactive and behavior-responsive (*23*), consistent with evidence that PA-based interventions yield larger improvements in anxiety than in depression (*24*). Alternatively, genetic predisposition may constrain the responsiveness of depression to active mobility more than anxiety, as suggested by our finding that genetic moderation, though modest overall, was relatively more apparent for depression. Previous studies have also linked active mobility with better mental health, but most focused on average associations or a single time point (*25, 26*), rather than estimating treatment effects, particularly at the individual level.

Turning to how these individual treatment effects were structured by urban contexts, we found that neighborhood environments strongly structure the mental health benefits of active mobility. Our composite measure of environmental supportiveness (THES) captures whether active mobility is undertaken in settings that are greener, cleaner, safer, and less deprived. This framing aligns with policy priorities on sustainable and healthy cities, equity, and safety. Mental health benefits were greater in more supportive environments with higher THES, consistent with evidence linking active commuting in natural environments to lower stress (*3, 5*). Spatially, enhanced effects were more common in rural areas where THES tended to be higher, whereas diminished effects clustered in major cities and other urban areas with lower THES.

Although compact-city strategies increase urban compactness to support accessibility and thus active mobility (*1, 2*), it remains unknown whether active mobility in compact urban form yields mental health co-benefits. Using our measure of functional urban compactness, we found that diminished health benefits clustered in major cities with higher urban compactness, such as Greater London, whereas enhanced benefits were more common in rural areas with lower compactness. Importantly, the associations between compactness and mental health effects of active mobility strongly depended on environmental supportiveness. Enhanced benefits were observed across compactness levels, but only in greener, cleaner, safer, and less deprived neighborhoods. In less supportive environments, benefits were consistently diminished and declined further as compactness increased. This may reflect higher exposure to traffic, noise, and safety-related stressors during active mobility in more compact settings, which can attenuate mental health gains (*30, 31*). Without buffering from supportive environmental quality, these stressors may outweigh the accessibility advantages of compactness (*32-34*). By contrast, greener, cleaner, and safer neighborhoods may mitigate these exposures, allowing accessibility advantages to translate into larger mental health gains (*35, 36*).

## Materials and Methods

### Study populations and area

Our study population were selected from the UKB, a nationwide population-based dataset with prospective follow-up. This dataset recruited around 500000 adults aged 39-70 at baseline across the UK between 2006 and 2010, with multiple ongoing follow-up assessments and linked electronic health records (*41*).

### Measurement of treatment

Transport mode for commuting was the treatment variable in this study. At baseline, participants were asked about their commuting mode and were allowed to choose multiple modes, namely private motor vehicle, walking, cycling, and public transit.

### Measurement of health outcomes

The health outcomes were symptom-based measures of probable depression and anxiety derived from mental health surveys at baseline and online follow-up, and broader common mental disorders (CMDs) based on clinical diagnoses from linked EHR.

### Measurement of neighborhood environments

We developed a neighborhood environmental supportiveness score capturing how well each participant's surrounding neighborhood supports the realization of psychological benefits from active mobility. Each variable was transformed into standardized deciles (1–10) and aligned so that higher values indicate greater supportiveness (green space retained its natural ordering; $PM_{2.5}$, $NO_2$, deprivation, and crime were reverse-coded). The five deciles were summed to generate an overall score between 5 (least supportive) and 50 (most supportive). We refer to this measure as transport-health environmental supportiveness (THES), hereafter environmental supportiveness.

### Measurement of compact urban form

Compact urban form incorporated four density metrics, namely public transport, street movement, retail, and residential, to capture the intensity and accessibility of urban activity. Each component was converted to standardized deciles and the four deciles were then combined into an overall score between 4 (lowest compactness) and 40 (highest compactness). Metric definitions and data sources are summarized in Supplementary Table S2 and described in previous literature (*54*).

### Measurement of genetic susceptibility

To evaluate whether genetic susceptibility to mental disorders modifies treatment effects of active mobility, we estimated polygenic risk scores (PRS) for anxiety, depression, and CMDs. PRS were derived based on the genetic architecture from established genome-wide association (GWAS) studies (*55, 56*).

## Statistical analysis

Overall, we estimated treatment effects for each health outcome by comparing potential risks under active mobility versus motorized travel at baseline for all participants, using models tailored to each outcome type and time scale. Specifically, we estimated (i) probability of prevalent depression or anxiety at baseline; (ii) probability of incident depression or anxiety at the 2016–2017 online follow-up; and (iii) the 10-year risk of clinically diagnosed CMDs from linked EHR. The study design and conceptual framework of this study are shown in Fig. 1.

## Prediction models for treatment effects

We estimated average treatment effects (ATE) and individual treatment effects (ITEs) of active mobility on health outcomes using a causal framework. Propensity score was generated using a stacked machine learning (ML) method, which combined multiple base learners to provide the optimal ensemble prediction (*62*). Five-fold cross-validation was used to fit these models. More model details are provided in *Supplementary Method 2*, *Fig. S1*.

Treatment effects (TEs) of active mobility were reported as relative risk (RR), derived for per individual as the ratio of predicted outcome risk under active mobility to the predicted risk under the counterfactual motorized travel. RR > 1 indicates that higher risks under active mobility than motorized travel (smaller health benefits), and RR < 1 indicates lower risks under active mobility (greater benefits).

## ITEs, neighborhood environments, and urban compactness

We characterized dose-response associations between ITEs of active mobility and environmental supportiveness by fitting restricted cubic spline regressions. Additionally, associations between each THES components and the primary outcomes from symptom-based follow-up were estimated.

We evaluated whether environmental supportiveness may modify the association of urban compactness with ITEs by fitting restricted cubic spline models including terms for THES, compactness, and their interaction.

## ITEs, neighborhood environments, and genetics

To assess the modification effects of individual genetic predisposition to mental disorders on the mental-health benefits of active mobility, we conducted three complementary analyses. First, we compared PRS between participants with higher versus lower ITEs (median split) and quantified the effect sizes using Cohen's d. Second, we examined whether the distribution of ITE quartiles varied across PRS strata (categorized into nine equal-sized groups). Third, we assessed effect modification by genetic risk by fitting models with THES, PRS group, and interaction between them to determine whether the ITE–THES association differed between high- and low-PRS participants. Details of the statistical tests are provided in Supplementary Method 3.

Preprint version.

## Acknowledgments

This work was conducted under the UK Biobank project 99489, using data provided from UKB and other data owners. We thank colleagues at the Wellcome Sanger Institute for providing their expert advice for this study.

**Data and materials availability**: UK Biobank data are available to approved researchers through a standard material transfer agreement. All data supporting the findings of this study are included in the main text and Supplementary Materials.